\def\arxivversion{}  
\newif\ifarxiv
\ifdefined\arxivversion\arxivtrue\else\arxivfalse\fi
\newif\ififac
\ifarxiv\ifacfalse\else\ifactrue\fi

\documentclass{ifacconf}

\usepackage{graphicx}      
\usepackage{natbib}        
\usepackage{xcolor}        
\usepackage{amsfonts}      
\usepackage{amsmath}       
\usepackage{booktabs}      
\usepackage{multirow}      
\usepackage{subcaption}    
\usepackage{algorithm}
\usepackage{algpseudocode}
\usepackage{mdframed}
\usepackage{amssymb}
\usepackage[normalem]{ulem}  
\usepackage{url}  

\makeatletter
\def\author@note@fmt{%
  \setcounter{footnote}{4}%
  \def\thefootnote{\fnsymbol{footnote}}
}

\@frontindent=\textwidth
\advance\@frontindent by -\@frontmatterwidth
\makeatother

\newcommand{\pn}{\mathcal{P}(n)}

\makeatletter
\ififac
\renewcommand\section{\@startsection{section}{1}{\z@}
 {3\p@ \@plus 1\p@ \@minus 1\p@}   
 {2\p@ \@plus 1\p@ \@minus 1\p@}   
 {\centering}}                     

 \renewcommand\subsection{\@startsection{subsection}{2}{\z@}
  {1\p@ \@plus 1\p@ \@minus 1\p@}  
  {1\p@ \@plus 1\p@ \@minus 1\p@}  
  {\itshape}}                      
\makeatother
\fi

\begin{document}
\begin{frontmatter}

\title{Situation-Aware Interactive MPC Switching for Autonomous Driving}


\author[1]{Shuhao Qi\thanksref{cofirst}}
\author[1]{Qiling Aori\thanksref{cofirst}}
\author[2]{Luyao Zhang}
\author[1]{Mircea Lazar}
\author[1]{Sofie Haesaert}

\address[1]{Eindhoven University of Technology, The Netherlands}
\address[2]{Delft University of Technology, The Netherlands}

\thanks[cofirst]{These authors contributed equally to this work.
\newline This work is supported by the European project SymAware under grant No. 101070802, the European project COVER under grant No. 101086228, and the Dutch NWO Veni project CODEC under grant No. 18244. (Corresponding author: {\tt\small s.qi@tue.nl})
\ififac \\
An extended version of this paper with full details and results is available at \protect\url{https://arxiv.org/abs/2512.06182}.
\fi}

\begin{abstract} 
Autonomous driving in interactive traffic scenarios remains challenging because of the mutual influence among vehicles and the inherent uncertainty of surrounding agents. Several model predictive control (MPC) formulations have been proposed to address this challenge, each adopting a different model of inter-agent interaction. While higher-fidelity interaction models enable more intelligent behavior, they incur substantially greater computational cost. Since strong interactions arise only occasionally in real traffic, a practical strategy for balancing performance and computational overhead is to invoke an appropriate controller based on situational demands. To this end, we first conduct a comparative study to assess and hierarchize the interactive capabilities of different MPC formulations. Building on this hierarchy, we then develop a neural network-based classifier for situation-aware switching among these controllers. We demonstrate that, by invoking the most advanced interactive MPC only in rare but critical situations and relying on a basic MPC in the majority of situations, situation-aware switching substantially improves overall performance while significantly reducing computational load.
\end{abstract}


\end{frontmatter}

\section{Introduction}


Advanced driver-assistance systems have reached mass deployment, yet fully autonomous driving still struggles in interactive scenarios such as lane merging and roundabouts, where human intervention is still frequently required. Following~\citep{Wang_2022}, an \emph{interaction} arises when two or more road users compete for the same space in the near future, so that their behaviors become mutually influential~\citep{icra2021game} and inherently multi-modal~\citep{chen2021branch}. Achieving safe and efficient driving in interactive scenarios under limited on-board computation remains an open problem.

Model predictive control (MPC) is well-suited to autonomous driving thanks to its native handling of constraints and uncertainty~\citep{MESBAH2018107}. Existing interactive MPC formulations differ primarily in how they model the opponent and form a hierarchy of increasing fidelity. \emph{Robust MPC} treats the opponent's future behavior as adversarial and computes control inputs against worst-case realizations~\citep{zhou2022robust}. The resulting policies are safe but conservative, since improbable worst cases dominate the optimization and realistic interaction patterns are disregarded. \emph{Branch MPC}~\citep{chen2021branch}, also known as \emph{scenario-based MPC}~\citep{9133136}, mitigates this conservatism by enumerating a scenario tree of plausible opponent trajectories~\citep{7990647, 9294176}. Early variants sample the opponent independently of the ego~\citep{9133136, 7990647, 9294176}, yielding non-reactive predictions and ignoring the mutual influence between the vehicles. More recent variants account for the opponent's reaction to the ego trajectory through a parametric driver model~\citep{chen2021branch, oliveira2023interactiondecisionmakingawaremotion}, frequently parameterized by latent driving styles or intents~\citep{tian2021anytime, hu2024active}. All of the above remain passive in uncertainty reduction: they wait for the opponent to reveal its intent. \emph{Dual MPC}~\citep{hu2024active} instead exploits the dual-control effect~\citep{knaup2024active, sadigh2018planning} by coupling estimation with control, selecting actions that simultaneously optimize task performance and actively reduce uncertainty. A notable interactive behavior emerges from this coupling: the controller probes the opponent to reveal its intent, then exploits the resulting information to improve behavior. In summary, these MPC controllers span three axes of progressively richer interactive behavior: multi-modal future, opponent reaction, and active information seeking.


Higher-fidelity interaction modeling, however, is not free: each step up the hierarchy raises computational cost. Since strong interactions are infrequent in everyday driving, applying the most advanced controller at every time step wastes onboard computation. We therefore propose \emph{situation-aware} switching: at each step, a classifier selects the lowest-fidelity controller sufficient for the current scenario. The contributions of this paper are as follows. First, we present a unified evaluation of five representative MPC formulations in interactive scenarios and empirically hierarchize their interactive capabilities through a Monte Carlo study. Second, we instantiate this switching mechanism with a neural-network classifier; the resulting policy preserves the interactive performance of the most advanced controller, such as active information seeking, while substantially reducing computation.


\section{Problem Formulation and Statement}\label{sec:problem}
We restrict the exposition to two-vehicle interactions; the formulation extends to multi-vehicle settings without conceptual change. We say that two vehicles are in an interactive scenario if their predefined paths intersect. Fig.~\ref{fig:ramp_merge} shows an exemplary one-lane ramp merging scenario with vehicle-to-vehicle interaction~\citep{knaup2024active}.

\begin{figure}[h]
    \centering
    \includegraphics[width=\ififac 0.42\textwidth \else 0.45\textwidth \fi]{Figures/ramp_policy.png}
    \caption{One-lane ramp merging. The ego car (red) merges into the lane of the opponent car (blue). The opponent has different predictions under cautious ($g_{\text{cau}}$) or aggressive behavior ($g_{\text{agg}}$) when close to the ego.}
    \label{fig:ramp_merge}
\end{figure}

The ego vehicle and opponent vehicle, denoted respectively with $i\in \{e,o\}$,  both follow a predefined path given as
\begin{align}
\label{eq:path}
\mathcal{T}^i:S\rightarrow \mathbb R^2 \textmd{ with } S\subset \mathbb R. 
	\end{align} 
 Along these paths, we define the longitudinal behavior \citep{9294176} based on  
 the  states  $x^i \!=\! [\,s^{i},\, v^{i}\,]^{\top} \!\in\! \mathbb{R}^2$  for vehicles $i\in\{e,o\}$, where $s^{i} \!\in\! S$ is the location on the path and $v^{i} \!\in\! \mathbb{R}$ is the longitudinal speed. The longitudinal dynamics of each vehicle are modeled by 
\begin{equation}
x^{i}_{t+1}  = f (x^{i}_{t}, u^{i}_{t})  
\label{eq:sys}
\end{equation}
\noindent where the input $u^{i}_{t}\!\in\!\mathbb{R}$ is the longitudinal acceleration. We consider interactive scenarios in which the paths $ \mathcal{T}^e$ and  $ \mathcal{T}^o$ intersect or overlap, that is, 
\begin{align}
\label{eq:intersect}
	\exists s^e, s^o: \|\mathcal{T}^e (s^e) - \mathcal{T}^o (s^o)\| < d_{\text{safe}},
\end{align}
where $d_{\text{safe}}$ denotes the safety distance between vehicles. Vehicles involved in such interactive scenarios are considered safe if they satisfy the following condition at all times:
\begin{equation}
	\forall t: h(x^{e}_{t}, x^{o}_{t}) := \|p^{e}_{t} - p^{o}_{t}\| - d_{\text{safe}} \geq 0, 
	\label{eq:safe_const}
\end{equation}  with the Cartesian position $p^{i}_{t} \!=\! \mathcal{T}^i(s^i_t) \!\in\! \mathbb{R}^2$.

\subsection{Opponent behavior model}\label{sec:oppo_model}

\begin{table*}[tb]
\centering
\caption{Comparison of MPC formulations in the ramp merging scenario.}
\vspace{-2mm}
\label{tab:overview}
\begin{tabular}{lcccc}
\toprule
MPC formulation & Multi-modal future & Opponent reaction &  Info-seeking & Related works \\
\midrule
Robust & $\times$ & $\times$ & $\times$  & \cite{langson2004robust,zhou2022robust} \\
Non-reactive branch  & $\checkmark$ & $\times$ & $\times$  & \cite{9294176,7990647} \\
Reactive branch & $\checkmark$ & $\checkmark$  & $\times$ & \cite{chen2021branch,oliveira2023interactiondecisionmakingawaremotion} \\
Explicit dual & $\checkmark$ & $\checkmark$ & $\checkmark$ & \cite{sadigh2018planning, baltussen2025online} \\
Implicit dual & $\checkmark$ & $\checkmark$ & $\checkmark$ & \cite{hu2024active,knaup2024active} \\
\bottomrule
\end{tabular}
\vspace{-2mm}
\end{table*}
The opponent's control input \(u^o_t\) is generated by a behavior model that emulates realistic driver behavior. Following driver models in~\citep{hu2024active, 9294176, wang2019learning}, we construct a parametric model that incorporates three key features: (i) uncertainty, (ii) responsiveness to other vehicles, and (iii) latent parameters. 
Accordingly, we model the opponent's control input at time $t$ as
\begin{equation}
\label{eq:oppo_policy}
u^o_t = g(x^o_t, x^e_t; \theta) + \epsilon_t, \quad \epsilon_t \sim \mathcal{N}(0, \sigma^2),
\end{equation}
where $g(x^o_t, x^e_t; \theta)$ is a parametric policy depending on the states of both vehicles, and $\epsilon_t$ is i.i.d.\ Gaussian noise capturing behavioral variability. The latent parameter $\theta \!\in\! \Theta$ is drawn from a finite set representing driver characteristics.
\ififac
A concrete example of this model, illustrated in Fig.~\ref{fig:ramp_merge}, is detailed in the extended version. This example adopts $\Theta \!=\! \{\theta_{\text{cau}}, \theta_{\text{agg}}\}$ representing cautious (yielding) and aggressive (non-yielding) behavior, with corresponding policies $g_{\text{cau}} \!:=\! g(\cdot;\theta_{\text{cau}})$ and $g_{\text{agg}} \!:=\! g(\cdot;\theta_{\text{agg}})$. The opponent tracks a reference speed when far from the ego and, as the two vehicles approach, switches to a reactive policy determined by the latent parameter.
\fi
\ifarxiv
\begin{mdframed}
\textbf{Example:} To model the likely reactions (yielding or not yielding) of an opponent vehicle as the ego vehicle approaches, we specify the policy $g(x^o_t, x^e_t; \theta)$ as the following piecewise expression,
\begin{equation}
\hskip -0.9em g(x^o_t, x^e_t; \theta) \!=\!
\begin{cases}
 K(v^{e} \!+\! \theta \!\cdot\! \Delta v \!-\! v^o), \hskip -0.8em & \|p^e_t - p^o_t\| \leq d_{\text{int}}\\
K(v^{o*} - v^o), & \text{otherwise}
\end{cases}
\label{eq:mean_func}
\end{equation}
\noindent where $v^o$ and $v^e$ are the velocities of the opponent and ego vehicles, respectively; $v^{o*}$ is a desired cruising speed; $\Delta v \!> \!0$ is a constant speed adjustment; $K$ is a control gain; and $d_{\text{int}}$ is a distance threshold below which the opponent initiates a reactive behavior.  The latent parameter $\theta\!\in\!\Theta$ is a scalar. For simplicity, we assume a binary set $\Theta \!=\! \{\theta_{\text{cau}}, \theta_{\text{agg}}\}$, with $\theta_{\text{cau}} \!=\! -1$ representing cautious (yielding) behavior and $\theta_{\text{agg}} \! = \! 1$ representing aggressive (non-yielding) behavior; we abbreviate the corresponding policies as $g_{\text{cau}} \!:=\! g(\cdot;\theta_{\text{cau}})$ and $g_{\text{agg}} \!:=\! g(\cdot;\theta_{\text{agg}})$. As shown in Fig.~\ref{fig:ramp_merge}, the opponent tracks $v^{o*}$ when far from the ego, and switches to one of two reactive policies within the interaction distance of the ego car.
\end{mdframed}
\fi

\subsection{Problem statement}

Given the interactive scenarios defined by~\eqref{eq:path} and~\eqref{eq:intersect}, the control objective is to track a desired ego speed $v^{e*}$ while satisfying the safety condition~\eqref{eq:safe_const} under an opponent governed by the behavior model~\eqref{eq:oppo_policy}. As outlined in the introduction, existing interactive MPC formulations (robust, branch, and dual MPC) model different aspects of the opponent and therefore exhibit different interactive behaviors and computational costs. We accordingly pursue two goals: first, to evaluate the interactive capability of five representative formulations under matched conditions; second, to design a situation-aware switching mechanism that selects, at each time step, the lowest-fidelity formulation sufficient for the current scenario.


\section{Interactive MPC Formulations}\label{sec:formulations}
We first formulate five representative MPC controllers in a unified framework that confines their differences to interaction modeling (multi-modal future, opponent reaction, and information seeking), as summarized in Table~\ref{tab:overview}.

\subsection{Robust MPC}
A common formulation of robust MPC is a min-max optimization problem~\citep{704989}, where the controller minimizes the worst-case cost to ensure constraint satisfaction. In interactive driving~\citep{zhou2022robust}, this worst-case cost is evaluated by having the opponent maximize the objective within its admissible set:
\begin{equation}
\begin{aligned}
& \min_{u^e_{0:H-1}} \max_{u^o_{0:H-1}} \sum_{k=0}^{H-1} \ell(x^e_k, u^e_k) + \ell_F(x^e_H) \hskip -1.5cm&& \\
& \text{s.t.} \quad x^e_{k+1} = f(x^e_k, u^e_k), && \hskip -0.2cm x^o_{k+1} = f(x^o_k, u^o_k),\\
& \phantom{\text{s.t.}} \quad h(x^e_{k+1}, x^o_{k+1}) \geq 0, && \hskip -0.2cm u^e_k \in \mathcal{U}^e, \quad u^o_k \in \mathcal{U}^o, \\
& \phantom{\text{s.t.}} \quad x^e_0,  x^o_0 \, \text{given}, && \hskip -0.2cm \forall k \in \{0, \dots, H-1\},
\end{aligned}
\label{eq:robust_mpc}
\end{equation} 
\noindent where $h(\cdot)$ is defined in~\eqref{eq:safe_const}. The terms $\ell(\cdot)$ and $\ell_F(\cdot)$ denote the stage cost and terminal cost, respectively. Here, the stage cost $\ell(x^e_k, u^e_k)$ penalizes deviation from a reference trajectory and the control effort, while the terminal cost $\ell_F(x^e_H)$ penalizes the final deviation.

\subsection{Branch MPC}
\label{sec:scenario MPC}
\begin{figure}[ht!]
    \centering
    \includegraphics[width=0.8\linewidth]{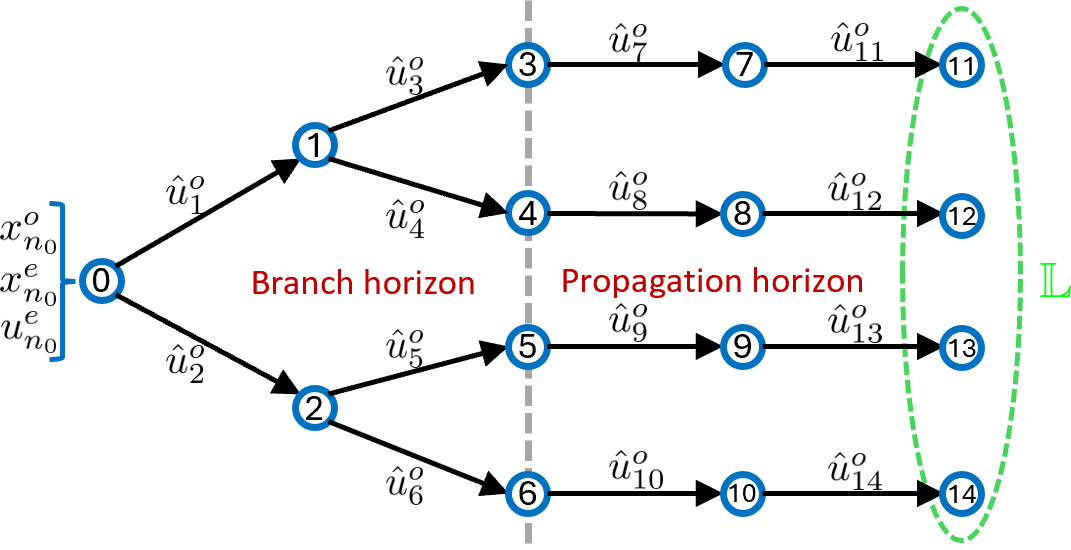} 
    \caption{A scenario tree with branching horizon $H_b\!=\!2$.}
    \label{fig:scen-tree-notation}
\end{figure}

Rather than optimizing for the worst case alone, a higher-fidelity controller plans over all plausible future interactions. To capture the multi-modal uncertainty of interaction, a \emph{scenario tree}~\citep{chen2021branch} is constructed by sampling possible opponent control inputs; each node represents one such sampled realization (Fig.~\ref{fig:scen-tree-notation}).

We use the following tree notation. Let $\mathbb{N}$ denote the set of $N$ tree nodes, with $n_0$ the (parent-less) root and $\mathbb{L}\!\subseteq\!\mathbb{N}$ the set of leaf nodes (those without children). For any non-root $n$, $\pn$ denotes its parent; for any non-leaf $n$, $\mathcal{C}(n)$ denotes its set of children. Each non-root node is associated with a sampled opponent control input $\hat{u}^o_n$. Given $\hat{u}^o_n$ and the ego control input at its parent node, $u^e_{\pn}$, the ego and opponent states at node $n$, denoted by $x^e_n$ and $x^o_n$, are determined according to the system dynamics specified in~\eqref{eq:sys}. Since each ego control input $u^e_n$ affects multiple child nodes, $u^e_n$ is optimized to account for multiple sampled realizations at its child nodes. To mitigate node explosion while maintaining a long prediction horizon, tree generation is split into two phases~\citep{hu2024active}: a branching horizon of length $H_b$ with multiple samples, followed by a propagation horizon with a single nominal sample, as shown in Fig.~\ref{fig:scen-tree-notation}.

\textbf{Non-reactive branch MPC:} Following a common constant-speed assumption~\citep{baltussen2025online}, we construct the scenario tree assuming the opponent maintains control inputs similar to its current behavior. As this assumption neglects potential reactions from opponent vehicles, it is referred to as~\emph{non-reactive branch MPC}. Specifically, at each tree node $n$, the opponent's control input is modeled as $\hat{u}^o_n \!=\! u_t^o + \epsilon_n$, where $u_t^o$ is the opponent's current control input and $\epsilon_n \!\sim\! \mathcal{N}(0,\sigma^2)$ is an independent disturbance realization at each node $n$. Let $\mathbf{U}^e \!=\! \{ u^e_n \}_{n \in \mathbb{N} \setminus \mathbb{L}}$ denote the set of ego control inputs at all non-leaf nodes, which are decision variables. Given this, the MPC minimizes the expected cost over the tree,
\begin{equation}
\begin{aligned}
& \min_{\mathbf{U}^e} \sum_{n \in \mathbb{N} \setminus \mathbb{L}} \ell \!\left(x^e_{n}, u^e_{n}\right)
+ \sum_{n \in \mathbb{L}}
\ell_F\!\left(x^e_{n}\right) \hskip -2cm && \\
& \text{s.t.} \quad x^e_{n} = f(x^e_{\pn}, u^e_{\pn}), &&  x^o_{n} = f(x^o_{\pn}, \hat{u}^o_{n}),\\
& \phantom{\text{s.t.}} \quad \hat{u}^o_n = u_t^o + \epsilon_n,  &&  h(x^e_n, x^o_n) \geq 0,  \\
& \phantom{\text{s.t.}} \quad u^e_{\pn} \in \mathcal{U}^e, &&  \forall n \in \mathbb{N} \setminus \{n_0\},
\end{aligned}
\label{eq:b_mpc}
\end{equation}
\noindent 
where the root node states $x^e_{n_0}$ and $x^o_{n_0}$ are initialized with the current actual states $x^e_t$ and $x^o_t$, respectively. The stage cost $\ell$ and terminal cost $\ell_F$ are identical to those in~\eqref{eq:robust_mpc}.

\textbf{Reactive branch MPC:} 
To incorporate plausible opponent reactions, the scenario tree is constructed using the opponent's behavioral model defined in~\eqref{eq:oppo_policy}. The opponent's control input at node $n$ is sampled from the behavior model evaluated at the parent node's state, with an additive disturbance realization:
\begin{equation}
    \hat{u}^o_n \!=\! g(x^o_{\pn}\!,\, x^e_{\pn};\theta_n) \!+\! \epsilon_n,
    \label{eq:react_sample}
\end{equation}
where $\epsilon_n \!\sim\! \mathcal{N}(0, \sigma^2)$ and $\theta_n \!\sim\! \text{Unif}(\Theta)$ are sampled offline at each node. Here, $\text{Unif}(\cdot)$ denotes a uniform distribution over a set. Replacing the non-reactive sampling $\hat{u}^o_n = u_t^o + \epsilon_n$ in \eqref{eq:b_mpc} with \eqref{eq:react_sample} yields the \emph{reactive branch MPC}.

\subsection{Dual MPC}

Similar to human drivers who can infer the driving styles and intentions of other vehicles through interaction, a higher-level controller should be capable of estimating the latent parameter $\theta$. Assuming the ego vehicle can perfectly observe the opponent's state, it has access to the information history at time $t$, denoted by $\mathcal{I}_t = \{ x^e_{0:t}, x^o_{0:t} \}$. Given the finite set of latent parameters $\Theta = \{\theta^{1}, \dots, \theta^{K}\}$ with $|\Theta| \!=\! K$, the belief state at time $t$ is defined as the posterior distribution over the latent parameters given the interaction history $\mathcal{I}_t$: $b_t = [b_t(\theta^1),\ldots,b_t(\theta^K)]^\top \in \mathbb{R}^K$, where $b_t(\theta^k)\!=\!\Pr(\theta\!=\!\theta^k \!\mid\! \mathcal{I}_t)$ and $\sum_{k=1}^{K} b_t(\theta^k) \!=\! 1$. As the interaction progresses, the ego vehicle observes the opponent’s next state $x^o_{t+1}$ and estimates the opponent’s input as $\tilde{u}^o_t \!=\! \frac{v^o_{t+1} - v^o_{t}}{\Delta t}$. Given $b_0$, the belief over latent parameters is updated via Bayesian inference:
\begin{equation}
b_{t}(\theta)\!=\! 
\dfrac{\rho_\theta (\tilde{u}^o_t \mid x^o_t, x^e_t)\, b_{t-1}(\theta)}
{\sum_{\theta' \in \Theta} \rho_{\theta'}(\tilde{u}^o_t \mid x^o_t, x^e_t)\, b_{t-1}(\theta')}, \, \forall \theta \in \Theta,
\label{eq:belief_update_function}
\end{equation}
where $\rho_\theta(\tilde{u}_t^o\!\mid\!x_t^o, x_t^e) = \tfrac{1}{\sqrt{2\pi\sigma^2}}\exp\!\big(\!-\!\tfrac{(\tilde{u}_t^o - g(x^o_t, x^e_t; \theta))^2}{2\sigma^2}\big)$ is the likelihood of observing $\tilde{u}^o_t$ under~\eqref{eq:oppo_policy}. This belief update is compactly expressed as $b_{t} \!=\! \mathcal{B}\big(b_{t-1}, x^o_{t}, x^e_{t}\big)$ using the operator $\mathcal{B}(\cdot)$. 



Since belief states quantify uncertainty and evolve with new information, incorporating belief updates into MPC formulations can induce a dual control effect. Two main approaches have been proposed~\citep{MESBAH2018107, arcari2020approximate}: the \emph{explicit} approach introduces an information gain term into the objective function to encourage probing behavior~\citep{sadigh2018planning}, whereas the \emph{implicit} approach incorporates belief evolution to couple estimation and planning~\citep{hu2024active}.


\textbf{Explicit dual MPC:} Building upon the reactive branch MPC framework, the \emph{explicit dual MPC} augments the objective function with an expected information gain associated with the current belief state:
\begin{equation}
\begin{aligned}
& \min_{\mathbf{U}^e} \sum_{n \in \mathbb{N} \setminus \mathbb{L}} \ell \!\left(x^e_{n}, u^e_{n}\right)
+ \sum_{n \in \mathbb{L}}
\ell_F\!\left(x^e_{n}\right)  \!+ \sum_{n \in \mathbb{N}} E(b_t, x^e_n, x^o_n) \hskip -6cm &&\\
& \text{s.t.} \  x^e_{n} \!=\! f(x^e_{\pn}, u^e_{\pn}), && \hskip -0.2cm x^o_{n} \!=\! f(x^o_{\pn}, \hat{u}^o_{n}),\\
& \phantom{\text{s.t.}}  \  \hat{u}^o_n \!=\! g(x^o_{\pn}\!,\, x^e_{\pn};\theta_n) \!+\! \epsilon_n, &&  \hskip -0.2cm h(x^e_n, x^o_n) \geq 0,\\
& \phantom{\text{s.t.}} \  u^e_{\pn} \!\in\! \mathcal{U}^e,  && \hskip -0.2cm \forall n \in \mathbb{N} \!\setminus\! \{n_0\},
\end{aligned}
\label{eq:ed_mpc}
\end{equation}
The information gain $E(b_t, x^e_n, x^o_n)$ is designed to elicit informative opponent reactions that facilitate belief estimation~\citep{sadigh2018planning, baltussen2025online}. \ififac An example of this gain function is provided in the extended version.\fi

\ifarxiv
\begin{mdframed}
\textbf{Example:} For the opponent model example given in~\eqref{eq:mean_func}, we define the gain function as
$$
E(b_t, x^e_n, x^o_n) = C_e \, b_t(\theta_{\text{cau}}) \, b_t(\theta_{\text{agg}}) \left((p_n^e - p_n^o)^2 - d_{\text{int}}^2 \right),$$
where $C_e$ is a constant weight, and the Cartesian positions $p_n^e$ and $p_n^o$ are computed according to~\eqref{eq:path}. The term $b_t(\theta_{\text{cau}}) b_t(\theta_{\text{agg}})$ serves as a measure of belief variance, which attains its maximum under a uniform belief distribution (i.e., $b_t = [0.5, 0.5]^\top$).
\end{mdframed}
\fi

\textbf{Implicit dual MPC:} 
The implicit approach does not include an explicit information gain and instead integrates belief updates over the tree. Specifically, the belief at node $n \!\in\! \mathbb{N}$ is denoted by $b_n$, with the root initialized at the current real-time belief, $b_{n_0} \!=\! b_t$. Each node's belief $b_n$ is computed recursively from its parent's belief $b_{\pn}$ using the state information at this node, $(x^e_n, x^o_n)$, with the belief update operator $\mathcal{B}(\cdot)$. The resulting implicit dual MPC is:
\begin{equation}
\begin{aligned}
& \min_{\mathbf{U}^e} \sum_{n \in \mathbb{N} \setminus \mathbb{L}} w_n \ell \!\left(x^e_{n}, u^e_{n}\right) +  \sum_{n \in \mathbb{L}}
w_n \ell_F\!\left(x^e_{n}\right) \hskip -6cm && \\
& \text{s.t.} \ x^e_{n} \!=\! f(x^e_{\pn}, u^e_{\pn}), && \hskip -0.2cm x^o_{n} \!=\! f(x^o_{\pn}, \hat{u}^o_{n}), \\
& \phantom{\text{s.t.}} \ \hat{u}^o_n \!=\! g(x^o_{\pn}\!,\, x^e_{\pn};\theta_n) \!+\! \epsilon_n, && \hskip -0.2cm h(x^e_n, x^o_n) \!\geq\! 0,  \\
& \phantom{\text{s.t.}} \ b_{n} \!=\! \mathcal{B}(b_{\pn}, x^o_{n}, x^e_{n}),  && \hskip -0.2cm u^e_{\pn} \in \mathcal{U}^e, \\
& \phantom{\text{s.t.}} \ x^e_{n_0},  x^o_{n_0} \, \text{given} && \hskip -0.2cm \forall n \in \mathbb{N} \setminus \{n_0\},
\end{aligned}
\label{eq:id_mpc}
\end{equation}
\noindent where the node weight $w_n$ represents the probability of reaching node $n$, assigning greater importance to nodes that are more likely to occur.
\ififac The belief update and node weighting are detailed in the extended version. \fi
\ifarxiv
At a non-root node $n$, the probability of a sampled $\hat{u}^o_n$ is obtained by marginalizing over the belief state of the parent node:
\begin{equation}
\rho(\hat{u}^o_n \mid \mathcal{I}_{\pn}) 
= \sum_{\theta \in \Theta} b_{\pn}(\theta) \, 
\rho_\theta(\hat{u}^o_n \mid x_n^o, x_n^e),
\label{eq:opponent_input_prob}
\end{equation}
where $b_{\pn}(\theta)$ denotes the belief over $\theta$ at the parent node, and $\rho_\theta(\hat{u}^o_n \!\mid\! x_n^o, x_n^e)$ represents the likelihood of sampled action $\hat{u}^o_n$ under the opponent policy parameterized by $\theta$. To normalize the sampled scenarios, the transition probability from the parent node $\pn$ to node $n$ is computed as
\begin{equation}
\Pr (n \mid \pn) 
= \frac{\rho(\hat{u}^o_n \mid \mathcal{I}_{\pn})}{
\sum_{i \in \mathcal{C}(\pn)} 
\rho(u_i^o \mid \mathcal{I}_{\pn})}.
\label{eq:transition_prob}
\end{equation}
The node weight is propagated recursively from the root:
\begin{equation}
w_n = w_{\pn} \, \Pr (n \mid \pn), 
\qquad
w_{n_0} = 1.
\label{eq:weight_recursion}
\end{equation}
Intuitively, this weighting downweights nodes whose sampled $\theta_n$ disagrees with the parent-node belief.
\fi

\begin{table*}[bt]
\centering
\caption{Monte Carlo simulations in the ramp-merging scenario.}
\vspace{-2mm}
\label{tab:comparison}
\scriptsize
\begin{tabular}{l|ccccc|c}
\toprule
 Metric / MPC formulations & Robust & Non-reactive branch & Reactive branch  & Explicit dual & Implicit dual & Switching\\
\midrule
Safety rate (\%) $\uparrow$   & 100.0 & 100.0 & 100.0 & 100.0 & 100.0 & 100.0 \\
Min distance (m) $\uparrow$  & \textbf{4.13 ± 0.29} & 1.99 ± 0.30 & 2.67 ± 0.27 & 3.11 ± 0.25 & 3.02 ± 0.27 & 3.24 ± 0.35 \\
\midrule
Front-merge rate (\%) $\uparrow$  & 1.0 & 4.6 & 24.6 & \textbf{47.3} & 43.0 & 44.8 \\
Completion time (s) $\downarrow$ & 5.85 ± 0.48 & 5.22 ± 0.42 & 4.93 ± 0.37 & \textbf{4.62 ± 0.32} & 4.65 ± 0.37  & 4.67 ± 0.21 \\
\midrule
Max abs. acc. (m/s²) $\downarrow$ & 3.39 ± 0.41 & 2.21 ± 0.56 & 1.02 ± 0.45 & 1.52 ± 0.53 & \textbf{0.93 ± 0.42} & 0.88  ± 0.47 \\
Control effort $\downarrow$    & 105.3 ± 7.8 & 54.1 ± 7.3 & 16.2 ± 4.5 & 22.9 ± 5.9 & \textbf{12.6 ± 3.7} & 14.5 ± 6.1 \\
\midrule
Trajectory cost $\downarrow$  & 578.8 ± 42.6 & 314.8 ± 45.7 & 109.7 ± 29.1 & 80.6 ± 23.5 & \textbf{68.9 ± 22.3} & 72.3 ± 19.8  \\
\midrule
Average computation (s) $\downarrow$ & \textbf{0.047 ± 0.005} &  0.324 ± 0.020  &  0.431 ± 0.115   &  0.434 ± 0.116  &  1.148 ± 0.148 &  0.313 ± 0.279 \\
\bottomrule
\end{tabular}
\end{table*}

\section{Controller Comparison and Switching Strategy}\label{sec:switching}
We next compare the five MPC controllers empirically and use the results to motivate a learned switching strategy.

\ifarxiv
\subsection{Implementation detail}
The vehicle dynamics in~\eqref{eq:sys} are realized as a double integrator, and the opponent follows the parametric model of~\eqref{eq:mean_func}. Following~\citep{Platt-RSS-10}, the scenario tree branches binarily on $g_{\text{cau}}, g_{\text{agg}}$. The ego vehicle begins with a uniform prior belief $b_0 = [0.5, 0.5]^\top$. The prediction horizon and branching horizon are set to $H = 15$ and $H_b = 2$, respectively, with a time step of $\Delta t = 0.05$. All formulations are implemented in Python using \texttt{CasADi}~\citep{Andersson2019} and solved using the interior-point optimizer \texttt{IPOPT}~\citep{wachter2006implementation}, executed on a laptop with an Intel Core i9-14900HX CPU.
\fi

\subsection{Comparative evaluation}

\ifarxiv
To compare the formulations fairly, shared parameters (e.g., objective weights) are identical across controllers, while controller-specific parameters are tuned to produce similar behavior in non-interactive simulations. Fixed random seeds ensure identical scenario initialization and opponent samples across controllers.
\fi
We evaluate controller performance through Monte Carlo simulations in the ramp merging scenario, with randomized initial positions and velocities centered around likely interaction scenarios.
\ififac
The vehicle dynamics in~\eqref{eq:sys} are realized as a double integrator; further implementation details are provided in the extended version.
\fi
The results of Monte Carlo simulations are summarized in Table~\ref{tab:comparison}, reported as mean~$\pm$~standard deviation over 600 runs. Each MPC formulation is tested on 300 trials per opponent type (aggressive and cautious). Performance is assessed using multiple metrics: safety is evaluated via collision rate and minimum inter-agent distance; efficiency by front-merge rate and completion time; comfort and energy by control effort (cumulative squared inputs) and maximal absolute acceleration; and overall performance by trajectory cost, defined as the accumulated stage cost $\ell$ over the trajectory.

In Table~\ref{tab:comparison}, robust MPC yields the largest minimum distance, longest completion time, and highest control effort, indicating strong safety guarantees at the expense of conservative and less comfortable behavior. In contrast, non-reactive branch MPC reduces completion time while enhancing comfort and saving energy. Incorporating opponent reactions significantly increases the front-merge rate. Compared to both branch MPC variants, the two dual MPC variants achieve shorter completion times while preserving a larger minimum distance, reflecting simultaneous improvements in safety and efficiency attributable to increased front-merge behavior. Compared to the explicit approach that may initiate active probing even when unnecessary, the implicit approach achieves the lowest trajectory cost and superior overall performance by triggering probing only when deemed beneficial. The front-merge rate serves as a proxy for decision-level efficiency. Across formulations, adding scenario trees, opponent-reaction modeling, and belief updates each raises this rate, indicating that higher-fidelity interaction modeling produces less conservative interactive behavior without compromising safety.

\subsection{Switching interactive MPCs as needed}

Table~\ref{tab:comparison} shows that higher-fidelity interaction modeling improves performance but increases computational time. Specifically, the computational burden of the implicit dual MPC is over 20 times greater than that of the robust MPC. Although more powerful hardware or more efficient algorithms can help accelerate computation, deploying dual MPC at every step would waste scarce onboard compute and energy. However, most driving situations do not require strong interaction with other agents. We therefore select the controller adaptively based on the driving situation.

\algrenewcommand{\algorithmicrequire}{\textbf{Input:}}
\begin{algorithm}[h]
  \caption{Situation-Aware Switching Controller}
  \label{alg:switch}
  \begin{algorithmic}[1]
    \Require Ego state $x_t^{e}$, opponent state $x_t^{o}$, belief state $b_t$
    \State Construct feature vector: $z_t \gets \left[(x_t^e)^\top, (x_t^o)^\top, b_t^\top\right]^\top$
    \State Compute controller probabilities: $\pi_t \gets \mathbf{NN}(z_t)$
    \State Select controller index: $i^\star \gets \arg\max_{i \in \{1,\cdots,m\}} \pi_t[i]$
    \State Compute control inputs:
    $
    \mathbf{U}_t^{e} \gets \text{MPC}_{i^\star}(x_t^e, x_t^o, b_t)
    $
    \State Extract the first control input: $u_t^e \gets \mathbf{U}_t^{e}[1]$
    \State \Return $u_t^{e}$
  \end{algorithmic}
\end{algorithm}
\vspace{-1mm}
Let $\text{MPC}_1, \ldots, \text{MPC}_m$ denote $m$ candidate controllers, ordered by increasing interactive capability. To enable situation-aware selection among them, we design a fully connected neural network $\mathbf{NN}: \mathbb{R}^d \to \mathbb{R}^m$ whose final layer is followed by a softmax, so that its output lies on the probability simplex. The network takes as input a feature vector representing the traffic scenario, defined as
$
z_t \!=\! \left[ (x^e_t)^\top, \!(x^o_t)^\top, \!b_t^\top \right]^\top \!\in\! \mathbb{R}^d,
$
which concatenates the ego state $x^e_t$, opponent state $x^o_t$, and belief state $b_t$. The network output is a probability vector $\pi_t \!=\! \mathbf{NN}(z_t)$, where $\pi_t[i]$ denotes the probability assigned to controller $\text{MPC}_i$. The switching protocol is provided in Alg.~\ref{alg:switch}. 


To construct the training dataset, we simulate over 10,000 randomly generated traffic scenarios. For each scenario with feature vector $z$, we score each controller offline using the composite performance metric:
$$
J(\text{MPC}_i; z) = J_{\text{traj}}(\text{MPC}_i; z) - \beta \cdot \mathbf{1}_{\text{front}}(\text{MPC}_i; z),
$$
where $J_{\text{traj}}(\text{MPC}_i; z)$ is the cumulative trajectory cost over the predictive horizon, $\beta \!>\! 0$ is a reward weight, and $\mathbf{1}_{\text{front}}(\text{MPC}_i; z)$ is an indicator function equal to 1 if controller $\text{MPC}_i$ selects a front-merge maneuver in scenario $z$, and 0 otherwise. The front-merge bonus is included because active probing incurs a short-term cost in $J_{\text{traj}}$ but may yield a shorter completion time in the long term by eliciting informative opponent reactions that enable a confident front-merge. Thus, $J_{\text{traj}}$ alone would discourage probing, and the bonus restores the long-term incentive. Let $J^*(z) \!=\! \min_i J(\text{MPC}_i; z)$ denote the best achievable cost in scenario $z$. A controller $\text{MPC}_i$ is considered \textit{near-optimal} if
$
J(\text{MPC}_i; z) \leq J^*(z) + \delta,
$
where $\delta \!>\! 0$ is a tolerance threshold. Each scenario is labeled with a one-hot vector $\pi^*(z)$, where only one entry is set to 1 and the rest to 0. The selected controller index is defined as
$
i^*(z) = \min \left\{ i \mid J(\text{MPC}_i; z) \leq J^*(z) + \delta \right\},
$
and the corresponding label is given by
$
\pi^*[i] \!=\! \mathbf{1}(i \!=\! i^*(z)),
$
favoring the lowest-level controller that achieves near-optimal performance. The tolerance $\delta$ governs how much cost the label is willing to forgo for a lower-fidelity controller: a larger $\delta$ resolves ties between adjacent capability levels in favor of the lower level, biasing the label toward cheaper controllers when the gain from a higher one is marginal.

\begin{figure*}[htbp]
    \centering
    
    \begin{subfigure}{\textwidth}
        \centering
        \includegraphics[width=0.925\textwidth]{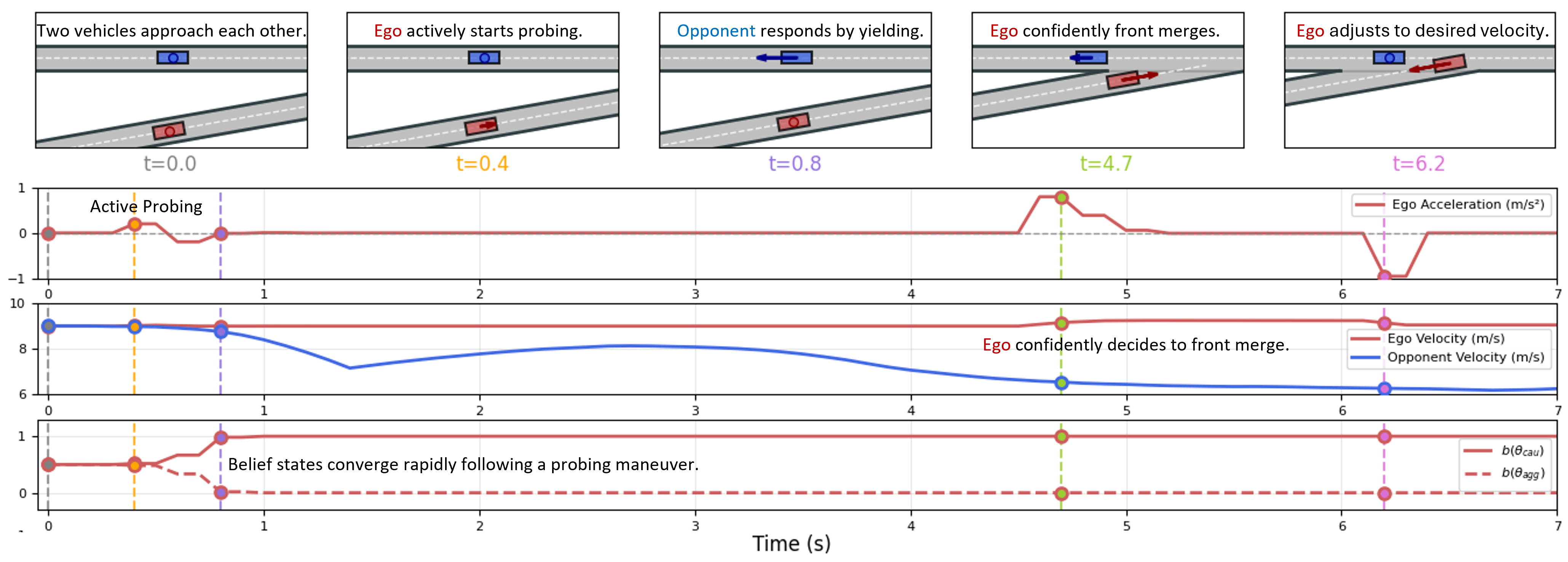}
        \vspace{-1em}
        \caption{Simulation of the implicit dual MPC interacting with a cautious opponent. The ego vehicle's desired velocity is set to 9 m/s. In the snapshots, the arrows on the vehicles represent their acceleration vectors.}
        \label{fig:probing_top}
    \end{subfigure}

    \vspace{0.5em}

    \begin{subfigure}{\textwidth}
        \centering
        \includegraphics[width=0.925\textwidth]{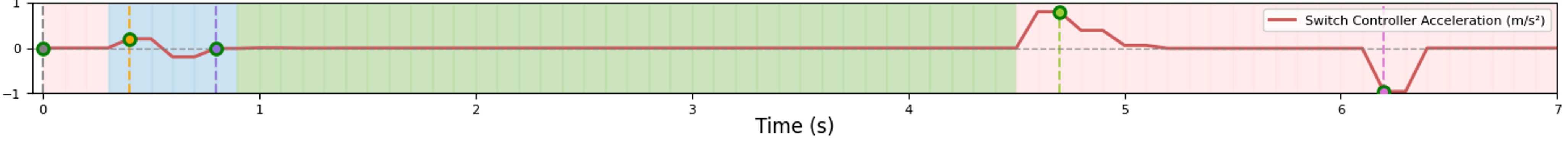}
        \vspace{-0.5em}
        \caption{Control inputs (accelerations) produced by the switching controller. The colored regions indicate which policy is active: blue for the implicit dual MPC, green for the reactive branch, and red for the robust MPC.}
        \label{fig:probing_bottom}
    \end{subfigure}
    \vspace{-2em}
    \centering
    \caption{Simulation of an interactive scenario that reveals active information-seeking (probing) behavior.}
    \label{fig:probing}
\end{figure*}

Because $i^*(z)$ is by construction the lowest near-optimal level, predicting $i \!<\! i^*(z)$ picks a controller whose cost exceeds $J^*(z) + \delta$, whereas predicting $i \!>\! i^*(z)$ only spends additional computation. Thus, we train $\mathbf{NN}$ with an asymmetric loss that penalizes incorrect choices of lower-level controllers more heavily than those of higher-level ones. Let $\pi_t = \mathbf{NN}(z_t)$ denote the predicted distribution over controllers and $\pi^*(z)$ the ground-truth label. The loss is defined as $\mathcal{L}(z) \!=\! \sum_{i=1}^{m} w(i, i^*(z)) \!\cdot\! \left( -\pi^*_i(z) \log \pi_i \right),
$ where $w(i, i^*)$ is a weighting function satisfying
$$
w(i, i^*) =
\begin{cases}
\alpha^{i^* - i}, & \text{if } i < i^*, \\
1, & \text{otherwise},
\end{cases}
$$
and $\alpha \!>\! 1$ is a tunable penalty factor that increases the cost of selecting a lower-level controller. Training the network with this loss yields a classifier that, at deployment, maps the current scenario feature $z_t$ to a controller index. The next section evaluates the resulting switching controller.

\section{Evaluation of Switching Controller}\label{sec:evaluation}
Among the five formulations of Section~\ref{sec:formulations}, we deploy three as candidates: robust MPC, reactive branch MPC, and implicit dual MPC, representing three levels of interactive capability.
\ifarxiv
The switching network is trained on 10{,}000 labeled scenarios using hyperparameters $\beta\!=\!10$ (front-merge reward), $\delta\!=\!1$ (near-optimality tolerance), and $\alpha\!=\!2$ (asymmetric-loss penalty factor). The neural network consists of three hidden layers with 128, 64, and 32 neurons, respectively. The trained neural network classifier achieves a test accuracy of 94.5\%.
\else
The trained neural network classifier achieves a test accuracy of 94.5\%; the training setup and network architecture are reported in the extended version.
\fi
We evaluate the switching controller within the same Monte Carlo simulation framework. The results, summarized in Table~\ref{tab:comparison}, show that the switching controller maintains performance comparable to that of the implicit dual MPC. The average computation time drops from $1.148$\,s for implicit dual MPC to $0.313$\,s for the switching controller, a reduction of roughly $73\%$ ($\sim\!3.7\times$). Across all simulation steps, only $1.4\%$ require the implicit dual MPC, while 42.3\% are handled by reactive branch MPC, and 56.3\% by robust MPC. This highlights a key result: invoking high-level interactive controllers in only a small subset of critical situations can lead to substantial performance improvements. This finding underscores the effectiveness of the switching strategy, which we further illustrate below\ifarxiv\ through two specific scenarios\fi.

\subsection{Ramp merging}
We select one representative ramp-merging scenario in which the dual MPC chooses a front-merge strategy, whereas other formulations opt for a back-merge. The entire process is illustrated in Fig.~\ref{fig:probing_top}. The acceleration profile reveals an active probing maneuver executed by the dual MPC at $t=0.6\,\mathrm{s}$. Specifically, the dual MPC applies a brief acceleration to elicit a response from the opponent vehicle. Upon observing a yielding reaction, the belief state rapidly converges, indicating a high likelihood that the driving style of the opponent is cautious. As a result, the ego vehicle confidently proceeds with the front-merge maneuver and maintains its desired speed, as shown in the velocity profiles. In contrast, the other approaches lack information-seeking behavior and default to a back-merge strategy due to parametric uncertainty.

This scenario clearly demonstrates the active probing behavior of the implicit dual MPC and highlights its benefits. We now examine whether the switching controller can inherit this advantage. To this end, we simulate the same scenario using the switching controller, which is also able to achieve a front-merge maneuver. The control inputs generated over time are shown in Fig.~\ref{fig:probing_bottom}. Notably, the implicit dual MPC is employed for only 8.6\% of the process, specifically when active probing is required. Once the belief state has converged, the switching controller transitions to a reactive branch MPC to maintain the desired speed before the merging point. As the ego vehicle approaches the merge location with a slight positional advantage for a front-merge, the controller switches to a robust MPC to ensure safety. This simulation demonstrates that the switching controller effectively replicates the active information-seeking behavior by invoking the dual MPC only in critical situations. \ififac More results on other scenarios are presented in the extended version.\fi

\ifarxiv
\subsection{Unsignalized intersection}

\begin{figure}[ht!]
\centering
\begin{subfigure}{0.47\linewidth}
    \centering
    \includegraphics[width=\linewidth]{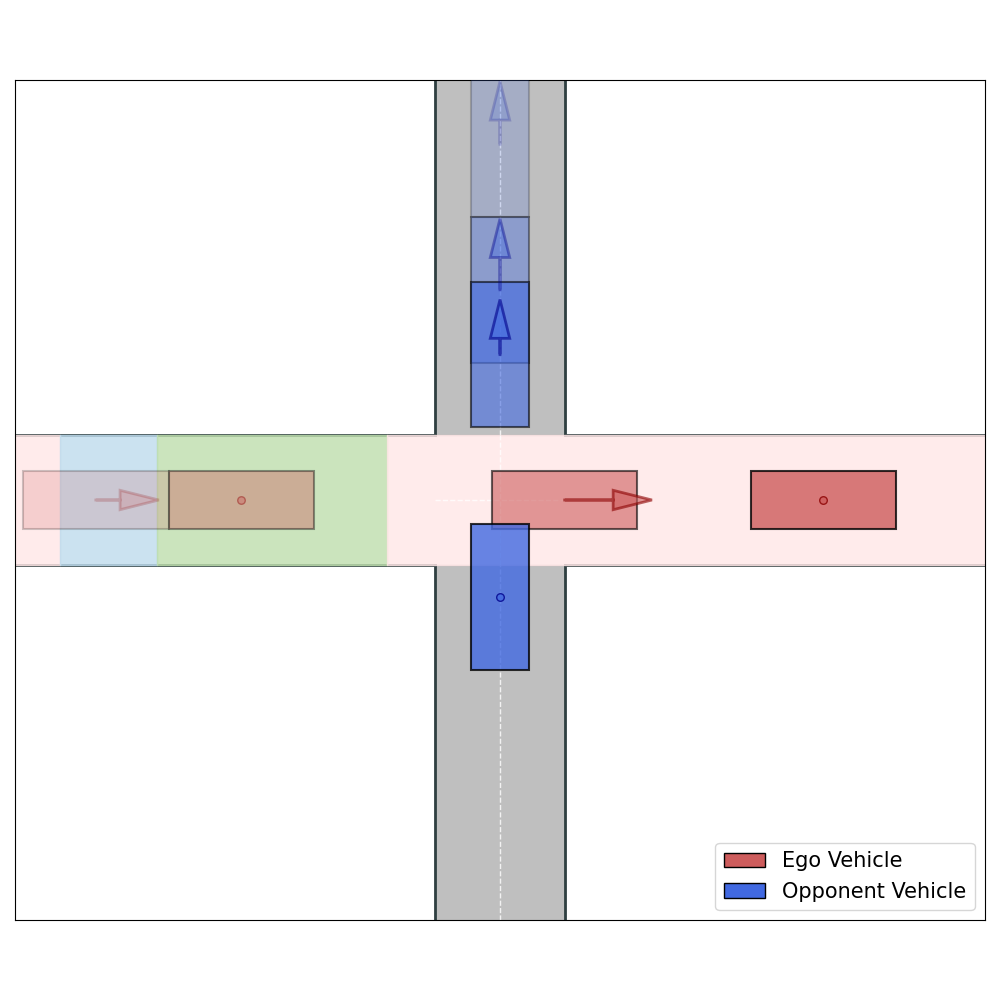}\vspace{-0.5em}
    \caption{Cautious opponent.}
    \label{fig:first}
\end{subfigure}
\hfill
\begin{subfigure}{0.47\linewidth}
    \centering
    \includegraphics[width=\linewidth]{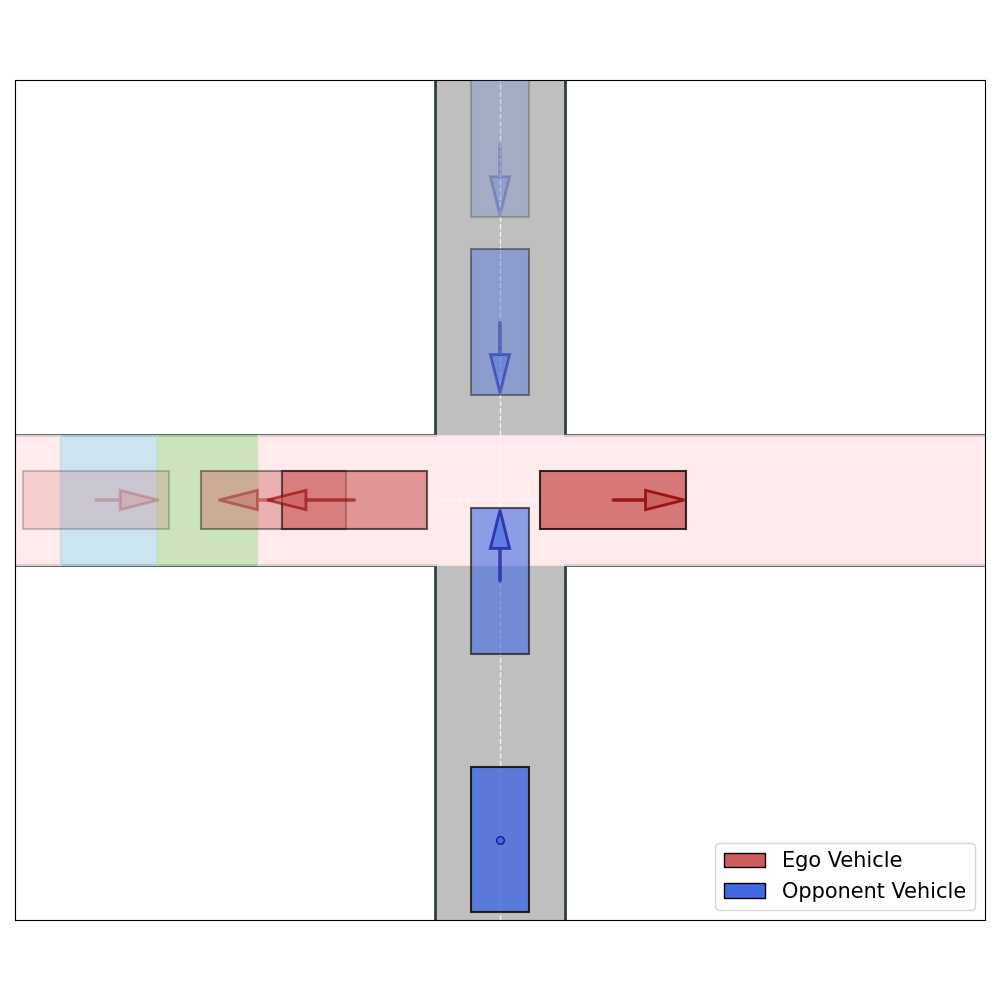}\vspace{-0.5em}
    \caption{Aggressive opponent}
    \label{fig:second}
\end{subfigure}
\vspace{-0.5em}
\caption{Interaction with an opponent in an unsignalized intersection. The colored regions indicate which policy is active: blue for the implicit dual MPC, green for the reactive branch, and red for the robust MPC.}
\label{fig:intersection}
\end{figure}

In Fig.~\ref{fig:intersection}, we evaluate the proposed method in an unsignalized intersection scenario. Similar to the ramp-merging case, the dual MPC is selected to trigger active probing behavior, while the robust MPC is employed to ensure safety as the vehicles approach each other. In Monte Carlo simulations with random initialization, the proposed switching controller achieves a 46.2\% first-crossing rate (the fraction of trials in which the ego vehicle clears the intersection ahead of the opponent, analogous to the front-merge rate of Sec.~\ref{sec:switching}), comparable to the 48.5\% of the implicit dual MPC and substantially higher than the 23.9\% of the reactive branch MPC and the 2.6\% of the robust MPC. Notably, its total computation time is reduced by 79.3\% relative to the implicit dual MPC and by 42.6\% relative to the reactive branch MPC. These results show that the approach generalizes beyond ramp merging to other interactive driving scenarios.
\fi

\section{Conclusion}\label{sec:conclusion}
This paper presented a unified comparison of representative MPC formulations for interactive driving and developed a learning-based switching mechanism that selects the lowest-fidelity controller sufficient for the current situation. Simulation results demonstrate that invoking high-fidelity controllers only in rare but critical situations substantially improves interactive performance while greatly reducing computational burden. Future work will extend this framework to more general interactive scenarios (e.g., multi-vehicle and combined longitudinal/lateral interactions) and establish theoretical guarantees for the resulting switching controller.



\bibliography{References}             
                                                   







\end{document}